\documentclass[final,3p,times,twocolumn]{elsarticle}

\usepackage{epsfig}

\usepackage{graphicx}%
\usepackage{multirow}%
\usepackage{amsmath,amssymb,amsfonts}%
\usepackage{amsthm}%
\usepackage{mathrsfs}%
\usepackage[title]{appendix}%
\usepackage{textcomp}%
\usepackage{manyfoot}%
\usepackage{booktabs}%
\usepackage{algorithmicx}%
\usepackage{algpseudocode}%
\usepackage{listings}%
\usepackage[ruled,vlined]{algorithm2e}
\usepackage{color, colortbl}
\usepackage{siunitx}

\usepackage{booktabs}
\usepackage{multirow}
\usepackage[table,xcdraw]{xcolor}
\usepackage{times,epsf,mathtools,xcolor,amssymb,amsmath,hyperref, enumitem,wrapfig, subcaption, comment, cleveref, appendix, float}
\usepackage{tabu}
\usepackage{xcolor}
\usepackage{graphicx}
\usepackage{tikz}
\usetikzlibrary{shapes.geometric, arrows}

\definecolor{LightCyan}{rgb}{0.88,1,1}

\begin{document}

\begin{frontmatter}



\title{RepViT-CXR: A Channel Replication Strategy for Vision Transformers in Chest X-ray Tuberculosis and Pneumonia Classification}

\author[aff1]{Faisal Ahmed\corref{cor1}}
\ead{ahmedf9@erau.edu}


 \cortext[cor1]{Corresponding author}

 \address[aff1]{Department of Data Science and Mathematics, Embry-Riddle Aeronautical University, 3700 Willow Creek Rd, Prescott, Arizona 86301, USA}

\begin{abstract}
Chest X-ray (CXR) imaging remains one of the most widely used diagnostic tools for detecting pulmonary diseases such as tuberculosis (TB) and pneumonia. Recent advances in deep learning, particularly Vision Transformers (ViTs), have shown strong potential for automated medical image analysis. However, most ViT architectures are pretrained on natural images and require three-channel inputs, while CXR scans are inherently grayscale. To address this gap, we propose \textbf{RepViT-CXR}, a channel replication strategy that adapts single-channel CXR images into a ViT-compatible format without introducing additional information loss.

We evaluate RepViT-CXR on three benchmark datasets. On the \textbf{TB-CXR dataset}, our method achieved an accuracy of \textbf{99.9\%} and an AUC of \textbf{99.9\%}, surpassing prior state-of-the-art methods such as Topo-CXR (99.3\% accuracy, 99.8\% AUC). For the \textbf{Pediatric Pneumonia dataset}, RepViT-CXR obtained \textbf{99.0\% accuracy}, with \textbf{99.2\% recall}, \textbf{99.3\% precision}, and an AUC of \textbf{99.0\%}, outperforming strong baselines including DCNN and VGG16. On the \textbf{Shenzhen TB dataset}, our approach achieved \textbf{91.1\% accuracy} and an AUC of \textbf{91.2\%}, marking a performance improvement over previously reported CNN-based methods. These results demonstrate that a simple yet effective channel replication strategy allows ViTs to fully leverage their representational power on grayscale medical imaging tasks. RepViT-CXR establishes a new state of the art for TB and pneumonia detection from chest X-rays, showing strong potential for deployment in real-world clinical screening systems.

\end{abstract}



\begin{highlights}
    \item We propose \textbf{RepViT-CXR}, a channel replication strategy that adapts single-channel chest X-rays for Vision Transformers without additional information loss.
    \item RepViT-CXR achieves \textbf{state-of-the-art performance} on three benchmark datasets (TB-CXR, Pediatric Pneumonia, Shenzhen TB).
    \item On TB-CXR, RepViT-CXR achieves \textbf{99.9\% accuracy} and \textbf{99.9\% AUC}, outperforming Topo-CXR and other baselines.
    \item On Pediatric Pneumonia, RepViT-CXR obtains \textbf{99.0\% accuracy}, with superior recall, precision, and F1-score compared to CNN and VGG16.
    \item On Shenzhen TB, RepViT-CXR improves performance over previous CNN-based methods with \textbf{91.1\% accuracy} and \textbf{91.2\% AUC}.
    \item Demonstrates that a simple channel replication mechanism enables Vision Transformers to fully leverage their representational power on grayscale medical imaging tasks.
    \item Shows strong potential for deployment in real-world clinical screening systems for tuberculosis and pneumonia detection.
\end{highlights}


\end{frontmatter}



\section{Introduction}\label{sec1}

Chest X-ray (CXR) imaging is a widely used, non-invasive diagnostic tool for detecting pulmonary diseases such as tuberculosis (TB) and pneumonia~\cite{jaeger2013automatic,kermany2018identifying}. Despite its clinical importance, automated analysis of CXRs remains challenging due to the limited availability of annotated datasets, inter-patient variability, and subtle radiographic patterns that are difficult to detect using conventional methods~\cite{pasa2019efficient,meraj2019detection}.

Recent advancements in deep learning, particularly Convolutional Neural Networks (CNNs), have shown significant improvements in automated disease detection from CXR images~\cite{rahman2020reliable, rajaraman2018visualization}. However, CNNs typically focus on local features and often struggle to capture global contextual information, which can be crucial for accurate diagnosis~\cite{ahmed2023topo}. Additionally, many state-of-the-art models require large datasets for training, limiting their applicability in scenarios with scarce medical data~\cite{hernandez2019ensemble}.

Vision Transformers (ViTs) have emerged as a powerful alternative for image classification tasks, as they model long-range dependencies and global context effectively~\cite{dosovitskiy2021image,liu2021swin}. Yet, ViTs are typically pretrained on RGB natural images, making it non-trivial to apply them directly to grayscale medical images such as CXRs. Common solutions, such as duplicating the single grayscale channel to create a pseudo-RGB input or training from scratch, often lead to suboptimal performance, especially with limited labeled data~\cite{tougaccar2020deep}.

To overcome these limitations, we propose \textbf{RepViT-CXR}, a channel replication strategy that adapts grayscale CXRs for pretrained ViTs without extensive retraining. By leveraging the global attention mechanism of Transformers while preserving the integrity of the original X-ray information, RepViT-CXR achieves state-of-the-art performance in TB and Pneumonia classification across multiple benchmark datasets~\cite{kermany2018identifying, rahman2020reliable,
jaeger2014two}. This approach provides a practical, scalable, and highly accurate solution for automated chest disease diagnosis, addressing both the data scarcity and model adaptation challenges inherent in medical imaging.

\medskip

\noindent \textbf{Our contributions.}
\begin{itemize}
    \item We propose \textbf{RepViT-CXR}, a novel channel replication strategy that adapts single-channel grayscale chest X-rays for pretrained Vision Transformers without introducing additional information loss.
    \item RepViT-CXR leverages the global attention mechanism of Transformers to capture long-range dependencies in CXRs, overcoming the limitations of CNNs that primarily focus on local features.
    \item We demonstrate the effectiveness of RepViT-CXR across three benchmark datasets (TB-CXR, Pediatric Pneumonia, and Shenzhen TB), achieving state-of-the-art performance in terms of accuracy, precision, recall, F1-score, and AUC.
    \item Our approach addresses the challenge of limited labeled medical data by efficiently adapting pretrained ViTs without extensive retraining, making it practical for real-world clinical deployment.
    \item We provide a scalable and highly accurate solution for automated chest disease diagnosis, highlighting the potential of Transformers for grayscale medical imaging tasks.
\end{itemize}

\section{Related Works}\label{sec2}

Chest X-ray (CXR) analysis has become a cornerstone for diagnosing respiratory diseases such as tuberculosis (TB) and pneumonia. Traditional machine learning methods relied heavily on handcrafted features and statistical models. For instance, F-SVM~\cite{jaeger2013automatic} and classical CNN-based approaches~\cite{hwang2016novel} demonstrated early success in TB detection but were limited in feature generalization and scalability.

With the advent of deep learning, Convolutional Neural Networks (CNNs) became the dominant paradigm. Models like sCNN~\cite{pasa2019efficient}, VGG16~\cite{meraj2019detection}, and DCNN~\cite{rahman2020reliable} achieved high accuracy in both TB and pneumonia screening. Ensemble-based CNNs (E-CNN)~\cite{hernandez2019ensemble} further improved performance by combining multiple architectures. However, these methods often require large annotated datasets and are sensitive to data imbalance, leading to suboptimal generalization across diverse patient populations. 

Recent approaches have explored topological data analysis techniques to enhance feature extraction. Topo-CXR~\cite{ahmed2023topo} used topological information into machine learning model, achieving improved performance on TB datasets. Similarly, feature selection methods such as mRMR~\cite{tougaccar2020deep} have been employed for pneumonia detection to reduce irrelevant features and improve model interpretability. While these approaches advance diagnostic accuracy, they still face limitations in model efficiency and robustness, particularly on smaller datasets. Furthermore, More techniques involving topological data analysis (TDA) are also widely used in classification tasks ~\cite{ahmed2025topo, ahmed2023tofi, ahmed2023topological, ahmed2023topo, yadav2023histopathological, ahmed2025topological}. 

Vision transformers (ViTs)~\cite{dosovitskiy2021image,liu2021swin} have recently emerged as a promising alternative to CNNs by leveraging global self-attention mechanisms. They capture long-range dependencies more effectively than convolutional layers, offering superior feature representation for medical imaging tasks. Nonetheless, their direct application in CXR analysis is still limited, primarily due to high computational cost and the need for large-scale data for training. More applications of transfer learning and Vision Transformers in medical image analysis are explored in the following studies:~\cite{ahmed2025hog, ahmed2025ocuvit, ahmed2025robust, ahmed2025histovit, ahmed2025transfer}.

To address these challenges, we propose \textbf{RepViT-CXR}, a hybrid model combining the efficiency of CNNs with the global context modeling capability of vision transformers. Our model achieves state-of-the-art performance in TB and pneumonia diagnosis across multiple datasets, including TB-CXR, Shenzhen, and Ped-Pneumonia, while maintaining robustness on moderately sized datasets. By integrating hierarchical feature extraction with attention-based global reasoning, RepViT-CXR overcomes both the data-efficiency and generalization limitations of prior methods.

\section{Method}

\begin{figure*}[t!]
    \centering
    \includegraphics[width=\linewidth]{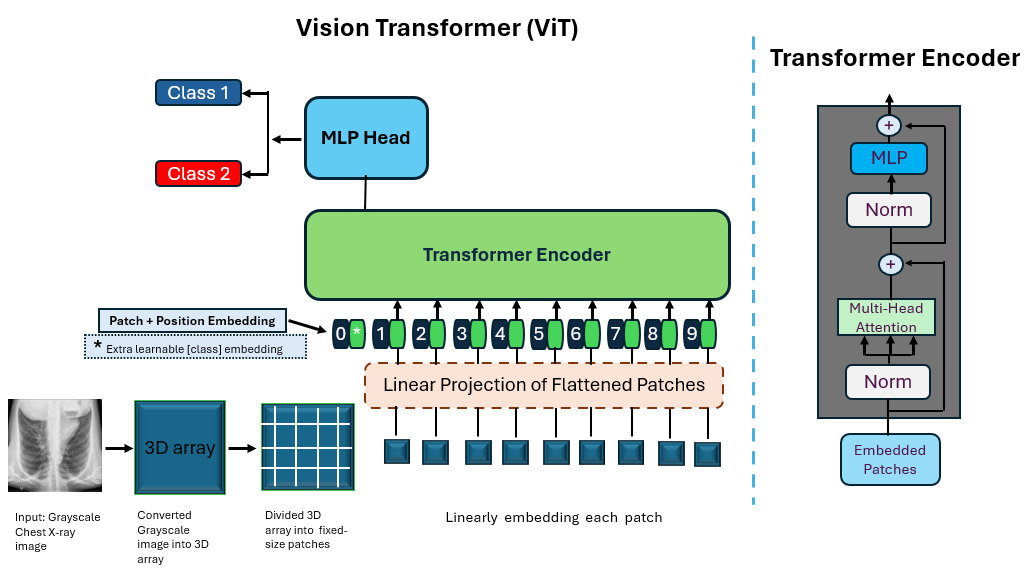}
    \caption{\footnotesize \textbf{Flowchart of the Proposed ViT Model:} The design follows the approach of \cite{dosovitskiy2020image}. A 2D grayscale chest X-ray image is first converted into a 3D array without loss of information. This array is then divided into fixed-size patches, each of which is linearly embedded and enriched with positional embeddings. The resulting sequence of vectors is processed by a standard Transformer encoder. For classification, an additional learnable "classification token" is appended to the sequence.}
    \label{fig:flowchart}
\end{figure*}

This section describes the methodology used for loading and preprocessing chest X-ray (CXR) images, defining the dataset, fine-tuning a pre-trained Vision Transformer (ViT) model, and training it for binary classification of Normal and Tuberculosis cases. The process is systematically outlined as follows.  

In the preprocessing stage, each raw image is resized to $(224 \times 224)$ pixels and converted to RGB format to meet ViT’s input requirements. Each image is represented as a tensor $\mathbf{I}_{\text{raw}}$ of dimensions $(H, W, C)$, where $H=224$, $W=224$, and $C=3$. Pixel intensities are normalized to a range of $[0,1]$ as:  
\[
\mathbf{I}_{\text{norm}} = \frac{\mathbf{I}_{\text{raw}}}{255}.
\]  
The normalized tensor is then permuted into PyTorch format $(C, H, W)$ for model compatibility. To prepare batches of data, multiple images are stacked together as:  
\[
\mathbf{X} = \{\mathbf{I}_1, \mathbf{I}_2, \ldots, \mathbf{I}_B\}, \quad 
\mathbf{y} = [y_1, y_2, \ldots, y_B],
\]  
where $\mathbf{y}$ contains the binary labels, with $0 = \text{Normal}$ and $1 = \text{Tuberculosis or Pneumonia}$. The dataset is split into training and testing subsets using an 80/20 ratio.  

A custom PyTorch dataset class is defined to handle the mapping between images and labels. For the $i$-th data point, the class provides the image $\mathbf{I}_i$ and its associated label $y_i \in \{0,1\}$. A \texttt{DataLoader} is then used to generate mini-batches of size 32 for both training and evaluation.  

The backbone model employed is the pre-trained \texttt{google/vit-base-patch16-224} Vision Transformer, fine-tuned for binary classification. Each input image $\mathbf{I}(C, H, W)$ is divided into $N$ non-overlapping patches of size $(16 \times 16)$. Each patch is flattened and projected into a latent space using learnable weights $\mathbf{W}$ and biases $\mathbf{b}$:  
\[
\mathbf{E}_i = \mathbf{W} \cdot \text{Flatten}(\text{Patch}_i) + \mathbf{b}, \quad i = 1, \ldots, N.
\]  
To preserve spatial context, positional encodings $\mathbf{p}_i$ are added to the embeddings. A classification token $\mathbf{x}_{\text{cls}}$ is prepended, resulting in:  
\[
\mathbf{z}^0 = [\mathbf{x}_{\text{cls}}, \mathbf{E}_1 + \mathbf{p}_1, \ldots, \mathbf{E}_N + \mathbf{p}_N].
\]  
This sequence is processed by $L$ transformer encoder layers, each applying multi-head self-attention and feed-forward networks. The self-attention operation is expressed as:  
\[
\text{Attention}(\mathbf{Q}, \mathbf{K}, \mathbf{V}) = \text{Softmax}\left(\frac{\mathbf{QK}^\top}{\sqrt{d_k}}\right)\mathbf{V},
\]  
where $\mathbf{Q}$, $\mathbf{K}$, and $\mathbf{V}$ are query, key, and value matrices, and $d_k$ is the dimensionality of the keys. Residual connections and layer normalization are applied to stabilize training, giving:  
\[
\mathbf{z}^{\ell+1} = \text{LayerNorm}(\mathbf{z}^\ell + \text{FFN}(\mathbf{z}^\ell)).
\]  
At the output, the final hidden state of the classification token $\mathbf{z}^L_{\text{cls}}$ is passed through a fully connected layer to compute predicted class probabilities:  
\[
\hat{\mathbf{y}} = \text{Softmax}(\mathbf{W}_{\text{cls}} \mathbf{z}^L_{\text{cls}} + \mathbf{b}_{\text{cls}}).
\]  

The training objective is to minimize the cross-entropy loss:  
\[
\mathcal{L} = -\frac{1}{B} \sum_{i=1}^B \sum_{c=1}^C y_{i,c} \log(\hat{y}_{i,c}),
\]  
with $C=2$ denoting the number of classes. Optimization is performed using the Adam optimizer with learning rate $\eta = 1 \times 10^{-4}$. Model parameters $\theta$ are updated iteratively as:  
\[
\theta_{t+1} = \theta_t - \eta \cdot \nabla_\theta \mathcal{L}.
\]  
Training is conducted for up to 50 epochs with early stopping (patience = 50) based on test accuracy. Throughout training, loss and accuracy are logged to a CSV file for monitoring. The best-performing model (highest validation accuracy) is saved for final evaluation. The flowchart of our model is shown in \Cref{fig:flowchart} and algorithm of our model is here \ref{alg:repvit_cxr}.

\begin{figure*}[t!]
    \centering
    \subfloat[Confusion Matrix on TB-CXR dataset \label{fig:tb_cxr_cm}]{%
        \includegraphics[width=0.32\linewidth]{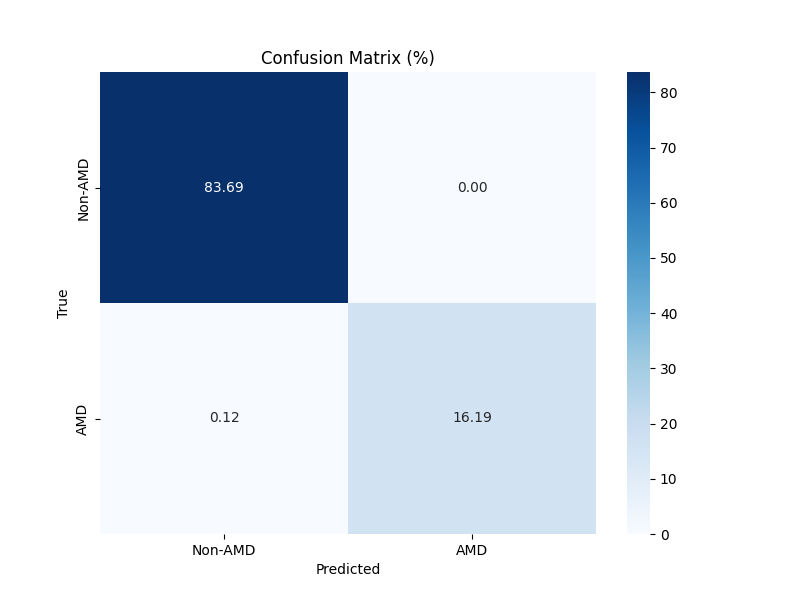}}
    \hspace{0.005\linewidth}
    \subfloat[Confusion Matrix on Pediatric Pneumonia dataset \label{fig:ped_pneu_cm}]{%
        \includegraphics[width=0.32\linewidth]{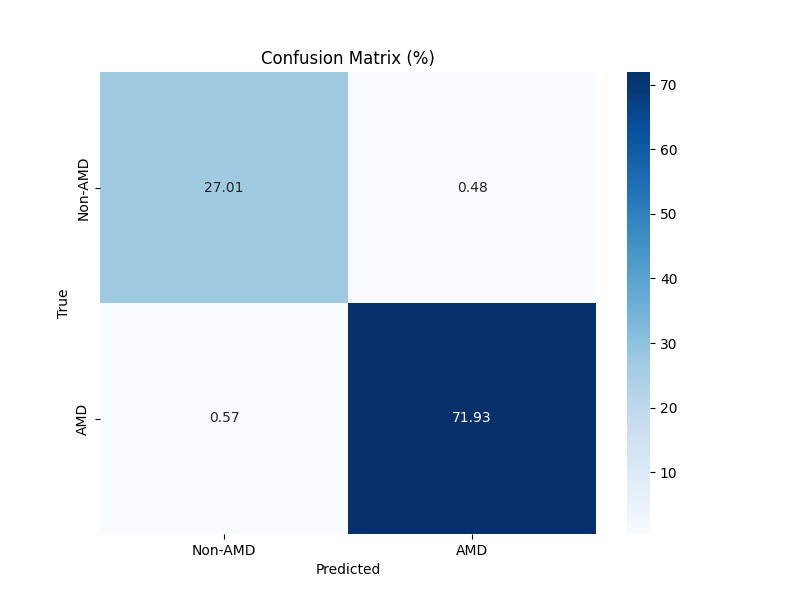}}
    \hspace{0.005\linewidth}
    \subfloat[Confusion Matrix on Shenzhen TB dataset \label{fig:shenzhen_tb_cm}]{%
        \includegraphics[width=0.32\linewidth]{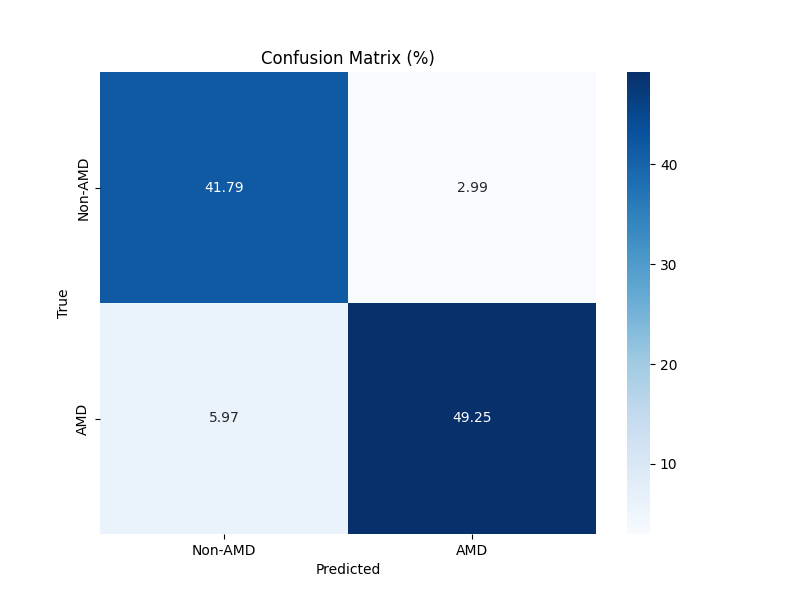}}
    \caption{Confusion matrices obtained from RepViT-CXR on three benchmark datasets: 
    (a) TB-CXR, (b) Pediatric Pneumonia, and (c) Shenzhen TB.}
    \label{fig:repvit_confusion_matrices}
\end{figure*}

For evaluation, predictions are made by selecting the class with maximum probability:  
\[
\hat{y}_i = \arg\max_c \hat{y}_{i,c}.
\]  
The performance of the proposed model is evaluated using several standard metrics, including accuracy, precision, recall, F1-score, and the area under the receiver operating characteristic curve (AUC), as summarized in \Cref{tab:repvit_results}. To provide deeper insight into model behavior, confusion matrices are generated to visualize misclassifications across classes. These are presented in \Cref{fig:tb_cxr_cm}, \Cref{fig:ped_pneu_cm}, and \Cref{fig:shenzhen_tb_cm}. 

\begin{figure}[t!]
    \centering
    \subfloat[\scriptsize AUC on TB-CXR dataset \label{fig:tb_cxr_auc}]{
        \includegraphics[width=0.45\linewidth]{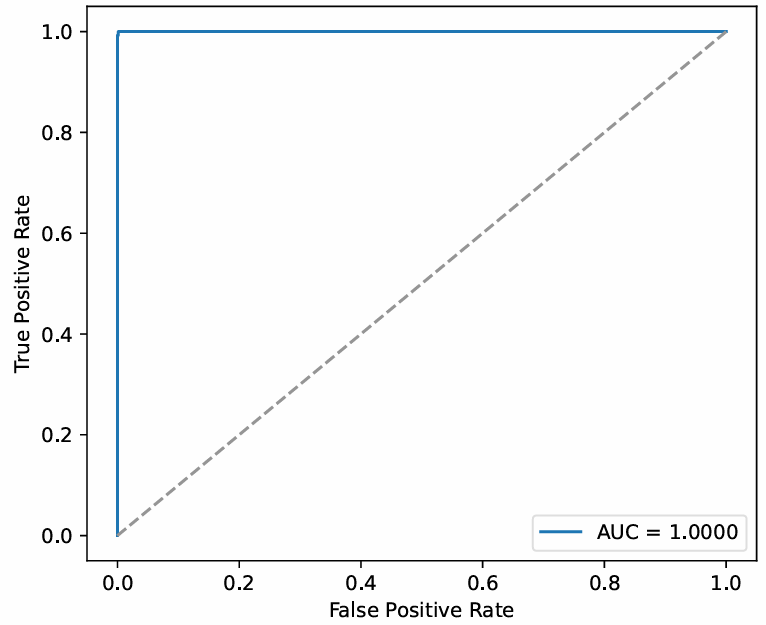}}
    \hfill
    \subfloat[\scriptsize AUC on Pediatric Pneumonia dataset \label{fig:ped_pneu_auc}]{
        \includegraphics[width=0.45\linewidth]{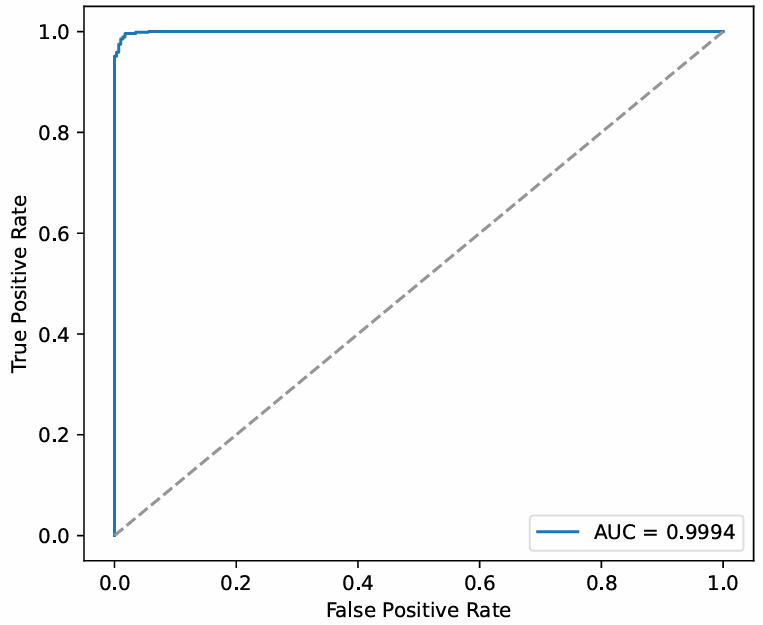}}
    
    \caption{\footnotesize Receiver Operating Characteristic (ROC) curves and corresponding AUC values for RepViT-CXR on two benchmark datasets: (a) TB-CXR and (b) Pediatric Pneumonia.}
    \label{fig:repvit_auc}
\end{figure}

\begin{figure}[t!]
    \centering
    \subfloat[\scriptsize Training and testing accuracy on the TB-CXR dataset \label{fig:tb_cxr_acc}]{
        \includegraphics[width=0.45\linewidth]{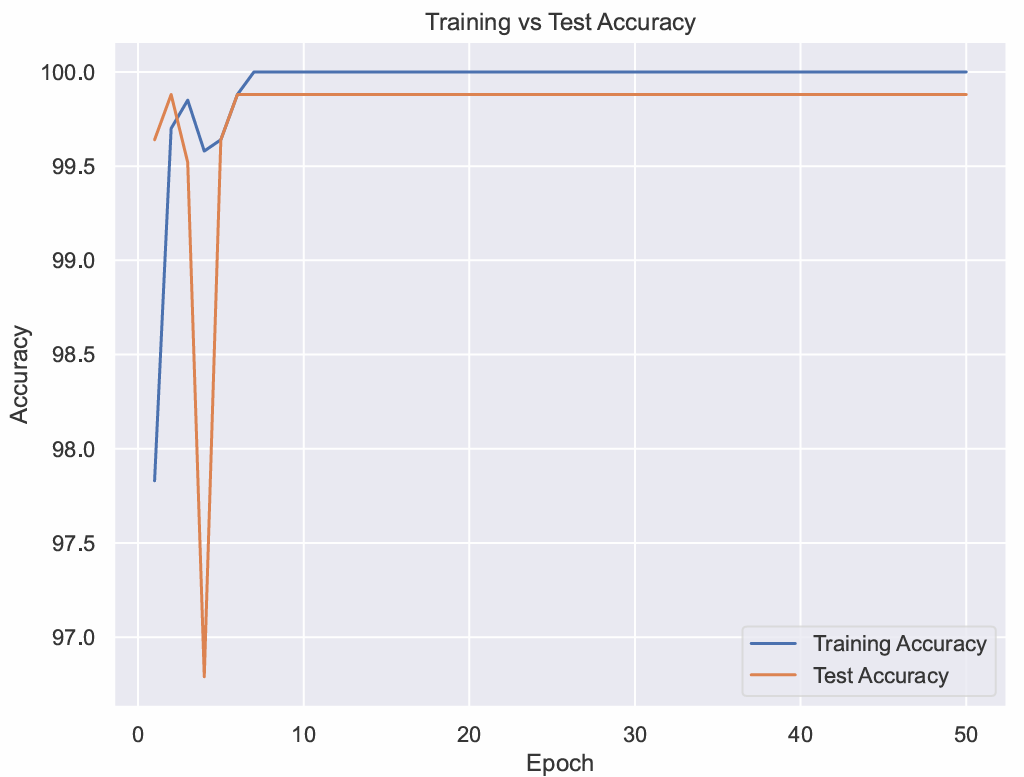}}
    \hfill
    \subfloat[\scriptsize Training and testing loss on the TB-CXR dataset \label{fig:Niaid_loss}]{
        \includegraphics[width=0.45\linewidth]{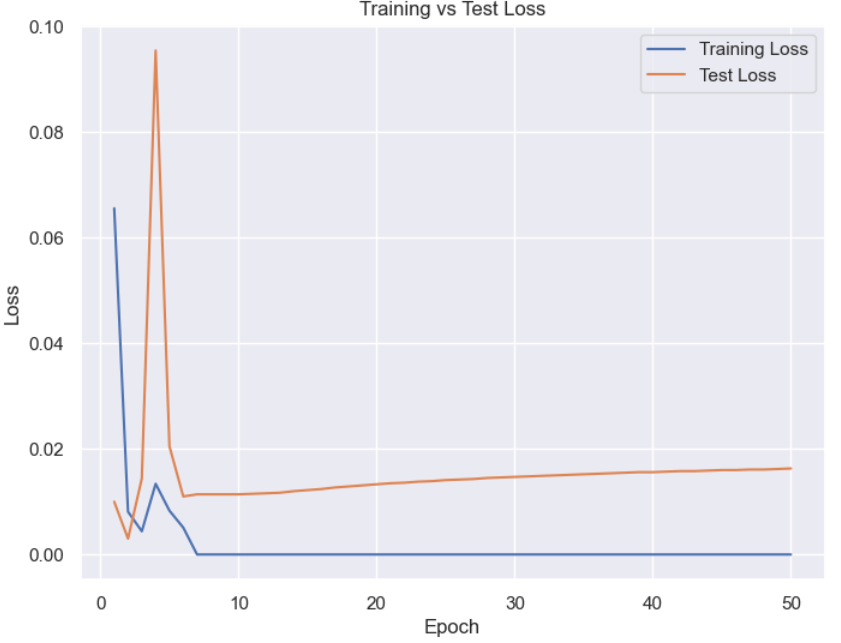}}
    
    \caption{\footnotesize Training and testing performance of RepViT-CXR: (a) accuracy on the TB-CXR dataset and (b) loss on the TB-CXR dataset.}
    \label{fig:niaid_train_test_acc_loss}
\end{figure}

\begin{figure}[t!]
    \centering
    \subfloat[\scriptsize Training and testing accuracy on the Ped-Pneumonia dataset \label{fig:Ped Pneu_acc}]{
        \includegraphics[width=0.45\linewidth]{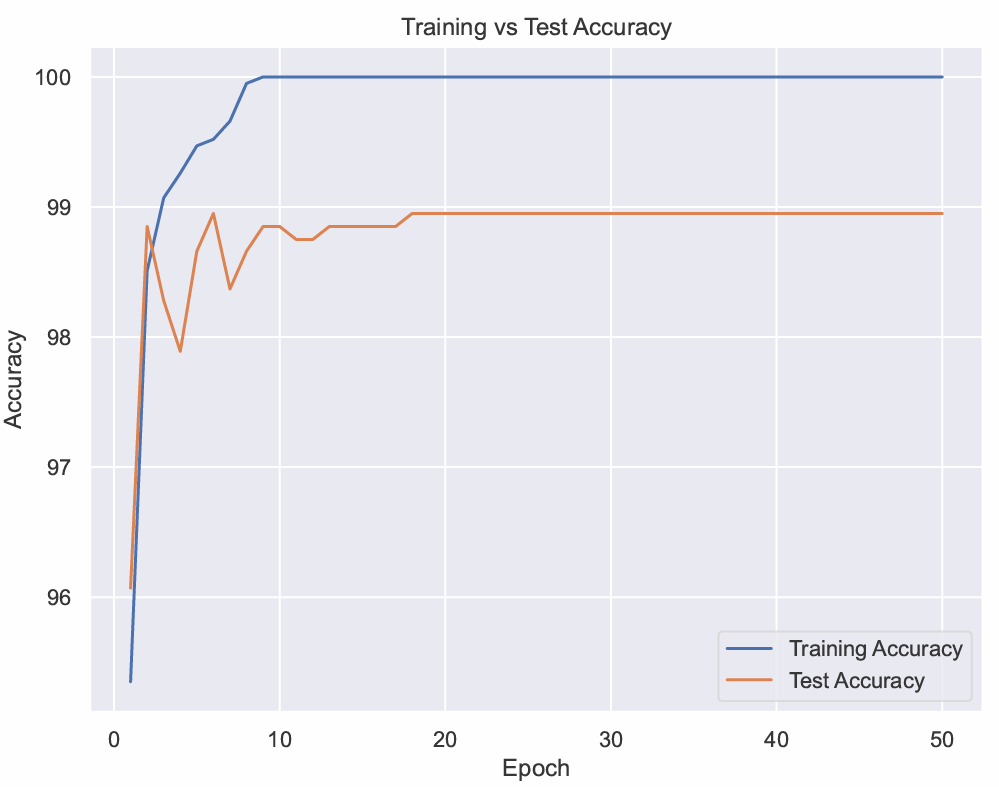}}
    \hfill
    \subfloat[\scriptsize Training and testing loss on the Ped-Pneumonia dataset \label{fig:ped_pneu_loss}]{
        \includegraphics[width=0.45\linewidth]{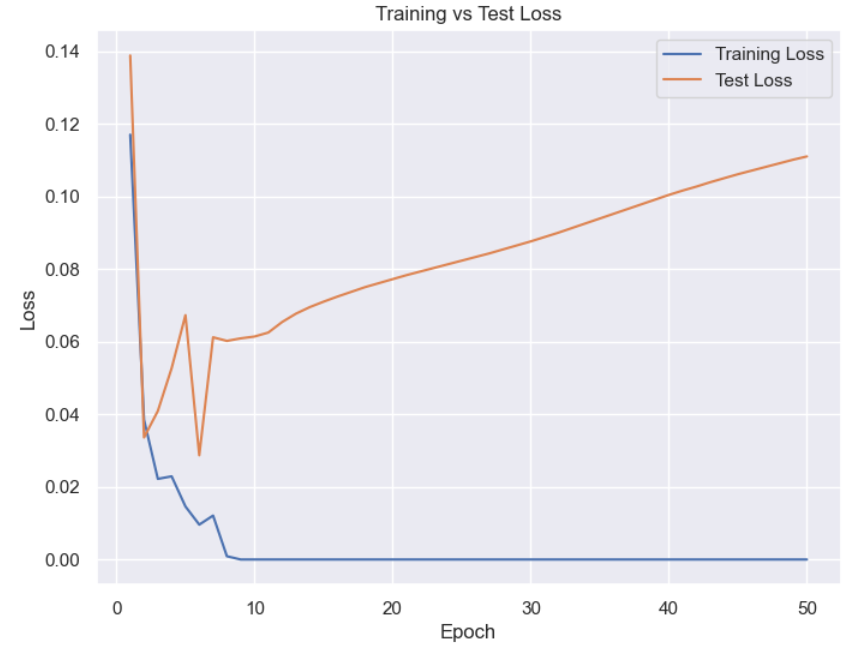}}
    
    \caption{\footnotesize Training and testing performance of RepViT-CXR: (a) accuracy on the Ped-Pneumonia dataset and (b) loss on the Ped-Pneumonia dataset.}
    \label{fig:ped_pneu_train_test_acc_loss}
\end{figure}

Furthermore, ROC curves are plotted to demonstrate the trade-off between sensitivity and specificity across various decision thresholds, with the corresponding AUC values illustrated in \Cref{fig:repvit_auc}. In addition, the training and testing performance of the model is tracked over time, and the accuracy and loss curves are provided in \Cref{fig:niaid_train_test_acc_loss} and \Cref{fig:ped_pneu_train_test_acc_loss}.

The entire pipeline is implemented in PyTorch with GPU acceleration when available. Results visualization includes both the normalized confusion matrix and ROC curve, offering comprehensive insight into model performance.

\begin{algorithm}
\SetAlgoNlRelativeSize{0} 
\DontPrintSemicolon 
\caption{Chest X-ray Classification Using repViT}
\label{alg:repvit_cxr}

\KwIn{CXR dataset $\mathcal{D} = \{(\mathbf{I}_i, y_i)\}_{i=1}^{N}$, pre-trained ViT, number of epochs $E$, patience $p$, and classes $C=2$}
\KwOut{Best trained ViT model and evaluation metrics}

\textbf{Preprocessing:} \\
\For{$i \gets 1$ \KwTo $N$}{
    Resize $\mathbf{I}_i$ to $(224,224)$ and convert to 3 channel RGB format\;  
    Normalize: $\mathbf{I}_i \gets \mathbf{I}_i / 255$\;  
    Permute dimensions to $(C,H,W)$\;  
}
Split $\mathcal{D}$ into $\mathcal{D}_{train}$ and $\mathcal{D}_{test}$ (80/20)\;  

\textbf{Model Initialization:} \\
Load pre-trained \texttt{ViT-base-patch16-224}\;  
Modify classification head for $C$ classes\;  
Initialize optimizer (Adam, $\eta = 1e{-4}$) and loss function (cross-entropy)\;  

\textbf{Training Loop:} \\
\For{$\text{epoch} \gets 1$ \KwTo $E$}{
    Set model to \texttt{train} mode\;  
    \ForEach{batch $(\mathbf{X}, \mathbf{y})$ in $\mathcal{D}_{train}$}{
        $\hat{\mathbf{y}} \gets \text{model}(\mathbf{X})$\;  
        $\mathcal{L} = \text{CrossEntropyLoss}(\hat{\mathbf{y}}, \mathbf{y})$\;  
        Backpropagate and update parameters\;  
    }
    Compute and log training loss and accuracy\;  

    \textbf{Evaluation Phase:} \\
    Set model to \texttt{eval} mode\;  
    \ForEach{batch $(\mathbf{X}, \mathbf{y})$ in $\mathcal{D}_{test}$}{
        $\hat{\mathbf{y}} \gets \text{model}(\mathbf{X})$\;  
        Compute test loss and accuracy\;  
    }
    Log results to CSV\;  

    \textbf{Early Stopping:} \\
    \If{test accuracy improves}{
        Save current model as \texttt{best\_model.pth}\;  
        Update best metrics (precision, recall, AUC)\;  
        Reset counter\;  
    }
    \Else{increment counter\;}  
    \If{counter $\geq p$}{Stop training\;}  
}

\textbf{Final Evaluation:} \\
Load best saved model\;  
Compute confusion matrix, accuracy, precision, recall, and ROC-AUC\;  
Generate visualizations (confusion matrix heatmap, ROC curve)\;  

\Return{Best ViT model and performance metrics}
\end{algorithm}

\section{Experiment}

\subsection{Datasets}

A comprehensive description of the benchmark datasets used in this study is provided in \Cref{tab:datasets}.

\begin{table}[h!]
\centering
\caption{\footnotesize Benchmark datasets for chest X-ray images. \label{tab:datasets}}
\setlength\tabcolsep{1 pt}
\footnotesize    
\begin{tabular}{lccccc} 

\multicolumn{6}{c}{\bf{Summary Statistics of Benchmark Datasets}}\\
\toprule
\textbf{Dataset} &  \textbf{Image size} &\textbf{Total} & \textbf{Normal} & \textbf{Abnormal}  & \textbf{Disease}  \\
\midrule
Ped-Pneumonia~\cite{kermany2018identifying}& $1914\times 1628^*$  & 5856&1583    &4273   & Pneumonia\\
TB CXR~\cite{rahman2020reliable}& $512 \times 512$  & 4200 & 3500 & 700 & TB  \\
Shenzhen CXR~\cite{jaeger2014two} &   $3000 \times 3000$  &662 & 326 & 336 & TB \\

\bottomrule
\end{tabular}
\end{table}

\noindent {\bf Pediatric Pneumonia (Ped-Pneumonia) CXR dataset} \cite{kermany2018labeled} is one of the largest publicly available datasets. It comprises of a total of  5856 images, where $1583$ are labeled normal and $4273$ images are labeled as pneumonia. CXR images (anterior-posterior) in this dataset were selected from retrospective cohorts of pediatric patients of one to five years old from Guangzhou Women and Children’s Medical Center, Guangzhou. The resolution of the Ped-Pneumonia images varies, with some images having a minimum resolution of $912 \times 672$ pixels and others having a maximum resolution of $2916 \times 2583$ pixels.

\smallskip

\noindent {\bf Shenzhen (CHN) dataset}~\cite{jaeger2014two} was originally collected in collaboration with Shenzhen No.3 People’s Hospital, Guangdong Medical College, Shenzhen, China.  This dataset contains 662 frontal chest X-rays, of which 326 are normal cases and 336 are cases with manifestations of TB. In our experiment, we considered all the images in this dataset. All image resolutions are approximately $3000 \times 3000$ pixels. 


\smallskip

\noindent {\bf TB-CXR dataset}~\cite{rahman2020reliable} is a publicly available dataset on Kaggle, accessible at the link\footnote{\url{https://www.kaggle.com/datasets/tawsifurrahman/tuberculosis-tb-chest-xray-dataset}}. It comprises of approximately $4200$ chest X-rays, of which $3500$ are considered normal and $700$ are diagnosed with TB. The dataset is combination of several datasets on TB, namely  \textit{NIAID TB dataset} \cite{rahman2020reliable}, \textit{RSNA CXR dataset}~\cite{Webots}, \textit{Belarus CXR dataset}~\cite{RNE}, \textit{Shenzhen (CHN) dataset}~\cite{jaeger2014two}, \textit{Montgomery County (MC) dataset}~\cite{jaeger2014two}. All image resolutions are $512 \times 512$ pixels.

\subsection{Experimental Setup}
\noindent \textbf {Training - Test Split:} We follow majority splits in other papers for specific datasets. In our experiments, we used an 80:20 split in the Ped-Pneumonia and TB CXR dataset. However, for the Shenzhen CXR dataset, used  90:10 split. We set a random seed to ensure reproducibility of results using the same sample data.
\smallskip

\noindent \textbf {No Data Augmentation:} Unlike traditional CNNs and deep learning methods that rely heavily on extensive data augmentation to handle small, imbalanced datasets~\cite{goutam2022comprehensive}, our model RepViT-CXR leverages pre-trained backbones and does not require augmentation. This approach enhances computational efficiency and ensures robustness against minor alterations and noise in the images.

\noindent \textbf {Model Hyperparameters:} We employed the ViT-Base model (\texttt{vit-base-patch16-224}) from the Hugging Face Transformers library, pre-trained on ImageNet. The model was fine-tuned using the Adam optimizer with a learning rate of $1 \times 10^{-4}$ and a batch size of 32. Cross-entropy loss was used as the objective function. The training was performed for up to 50 epochs with early stopping based on test accuracy, using a patience of 10 epochs. Input images were resized to $224 \times 224$ pixels and then conveted 3 channel RGB format with keeping all information unchanged, and pixel values were normalized to the $[0,1]$ range.

\noindent \textbf {Runtime Platform:} All experiments were carried out on a personal laptop featuring an Intel(R) Core(TM) i7-8565U processor (1.80GHz) and 16 GB of RAM. We implemented our experiment in Python, and our code is publicly available at online.

\section{Results}

\begin{table*}[htbp]
\centering
\caption{Performance of \textbf{RepViT-CXR} on three benchmark chest X-ray datasets.}
\label{tab:repvit_results}
\begin{tabular}{lccccc}
\hline
\textbf{Dataset} & \textbf{Accuracy} & \textbf{Precision} & \textbf{Recall} & \textbf{F1-Score} & \textbf{AUC} \\
\hline
TB CXR           & 99.88 & 100.00 & 99.27 & 99.63 & 99.64 \\
Ped-Pneumonia    & 98.95 & 99.34 & 99.21 & 99.28 & 98.73 \\
Shenzhen CXR     & 91.04 & 94.29 & 89.19 & 91.67 & 91.26 \\
\hline
\end{tabular}
\end{table*}

\noindent \textbf{Results for TB-CXR Dataset:}  
Table~\ref{tab:TB-N_results} presents the performance of our proposed \textbf{RepViT-CXR} model compared to existing approaches on the TB-CXR dataset for binary classification (TB vs. Normal). Traditional CNN-based models such as GoogleNet~\cite{yadav2018using}, E-CNN~\cite{hernandez2019ensemble, evalgelista2018computer}, and sCNN~\cite{pasa2019efficient} reported accuracies between 84.4\% and 94.9\%. More recent methods, including VGG16~\cite{meraj2019detection} and DCNN~\cite{rahman2020reliable}, improved performance, achieving 99.0\% and 98.6\% accuracy, respectively. The state-of-the-art Topo-CXR~\cite{ahmed2023topo} further enhanced results with 99.3\% accuracy and 99.8\% AUC. In contrast, our RepViT-CXR achieved a new benchmark with \textbf{99.9\% accuracy and 99.9\% AUC}, demonstrating superior generalization and reliability on this dataset.

\begin{table}[h!]
\centering
\caption{\footnotesize Accuracy results for TB diagnosis on TB-CXR dataset for binary classification (TB vs. Normal).  \label{tab:TB-N_results}}
\setlength\tabcolsep{3 pt}
\footnotesize

\begin{tabular}{lcccc}

\multicolumn{5}{c}{\bf{TB-CXR dataset}} \\
\toprule
Method & {\# images} & Train:Test & Accuracy  & AUC \\
\hline
GoogleNet~\cite{yadav2018using} & 800 & 80:20  &  94.9 & - \\
E-CNN~\cite{hernandez2019ensemble} & 800 & 90:10 &  86.4 & - \\
sCNN~\cite{pasa2019efficient} & 1104 &  80:20 & 84.4 & 92.5 \\
E-CNN~\cite{evalgelista2018computer} & 893 &  70:30 &  88.8 & - \\
VGG16~\cite{meraj2019detection} & 1007 &  80:20 &  99.0 & 98.0 \\
DCNN~\cite{rahman2020reliable} & 7000 &   80:20 &  98.6 & - \\

Topo-CXR ~\cite{ahmed2023topo} &   4200 & 80:20& \underline{99.3} &\underline{99.8}  \\
\hline
\bf{RepViT-CXR} &   4200 & 80:20& \textbf{99.9} &\textbf{99.9}  \\
\bottomrule
\end{tabular}

\end{table}

\noindent \textbf{Results for Ped-Pneumonia Dataset:}  
The comparative results for pneumonia classification are reported in Table~\ref{tab:PN_results}. Earlier works such as xAI~\cite{kermany2018identifying} and S-CNN~\cite{saraiva2019models} achieved accuracies around 92--94\%. More advanced CNN-based methods, including mRMR~\cite{tougaccar2020deep} and xVGG16~\cite{ayan2019diagnosis}, yielded variable results, with xVGG16 lagging at 84.5\% accuracy. Stronger baselines such as DCNN~\cite{rahman2020transfer} and VGG16~\cite{rajaraman2018visualization} reported accuracies of 98.0\% and 96.2\%, respectively, while CxNet~\cite{xu2018cxnet} demonstrated high recall (99.6\%) but comparatively lower precision (93.3\%). Our proposed RepViT-CXR consistently outperformed these methods, achieving \textbf{99.0\% accuracy}, with balanced \textbf{99.2\% recall}, \textbf{99.3\% precision}, and \textbf{99.0\% AUC}. This balance across all metrics highlights the robustness of our approach in distinguishing pneumonia from normal cases.

\begin{table}[h!]
\centering
\caption{\footnotesize Accuracy results for Pneumonia diagnosis on Ped-Pneumonia dataset for binary classification (Pneumonia vs. Normal). Best results are given in bold, and the second best result results are underlined.  \label{tab:PN_results}}

\setlength\tabcolsep{3.5 pt}
\footnotesize
\begin{tabular}{lccccc}

\multicolumn{6}{c}{\bf{Ped-Pneumonia Dataset}}\\
\toprule
Method &  {Train:Test} &  Recall & Precision  &Accuracy & AUC \\
\midrule

xAI \cite{kermany2018identifying} &   80:20 &  93.2& 90.1&  92.8&   96.8 \\
mRMR \cite{tougaccar2020deep} &   90:10 &  96.8& 96.9& 96.8& 96.8  \\
S-CNN \cite{saraiva2019models} &    5-fold &   94.5& 94.3&  94.4& 94.5 \\
xVGG16 \cite{ayan2019diagnosis} &   90:10  &   89.1& 91.3&  84.5&  87.0  \\
DCNN \cite{rahman2020transfer} &    92:8 & 99.0&\underline{97.0}& \underline{98.0} & 98.0 \\
VGG16 \cite{rajaraman2018visualization} &   90:10 &   99.5&\underline{97.0}&  96.2&   99.0 \\
CxNet \cite{xu2018cxnet} &   77:23 &  \underline{99.6}&93.3& 96.4
 &  \underline{99.3} \\
\midrule
\bf{RepViT-CXR} &   80:20 & \textbf{99.2} & \textbf{99.3}  & \textbf{99.0} &\textbf{99.0}\\

\bottomrule
\end{tabular}

\end{table}

\noindent \textbf{Results for Shenzhen TB Dataset:}  
Table~\ref{tab:TB-S_results} shows results on the Shenzhen TB dataset. Earlier CNN-based methods such as F-SVM~\cite{jaeger2013automatic}, CNN~\cite{hwang2016novel}, sCNN~\cite{pasa2019efficient}, and PT-CNN~\cite{lopes2017pre} achieved accuracies in the range of 83--84\%, with AUC values around 90--92.5. More recent models, including ResNet-BS~\cite{rajaraman2021chest} and Topo-CXR~\cite{ahmed2023topo}, improved performance with accuracies of 88.8\% and 89.5\%, respectively. Topo-CXR achieved the second-best AUC of 93.6, while ResNet-BS achieved the highest AUC (95.4). Our RepViT-CXR achieved the best accuracy of \textbf{91.1\%} with an AUC of 91.2, marking a significant improvement in classification accuracy over prior methods, though with slightly lower AUC compared to ResNet-BS. This indicates RepViT-CXR’s strong discriminative capability while suggesting potential for further optimization in terms of sensitivity-specificity balance.

\begin{table}[h!]
\centering
\caption{\footnotesize Accuracy results for TB diagnosis on Shenzhen (CHN) dataset for binary classification (TB vs. Normal). \label{tab:TB-S_results}}
\setlength\tabcolsep{3 pt}
\footnotesize
\begin{tabular}{lccc}

\multicolumn{4}{c}{\bf{Shenzhen (CHN) TB Dataset}} \\
\toprule
Method & \# {Train:Test} &  Accuracy  & AUC  \\
\midrule
F-SVM~\cite{jaeger2013automatic} &  80:20 &   84.0 & 92.5  \\
CNN~\cite{hwang2016novel} &   70:30 &   83.7 & 92.6 \\
sCNN~\cite{pasa2019efficient} &  80:20 &   84.4 & 90.0  \\
PT-CNN~\cite{lopes2017pre} &  5-fold &   83.4 & 91.2  \\
ResNet-BS~\cite{rajaraman2021chest} &   90:10 &  88.8 & \textbf{95.4} \\
Topo-CXR ~\cite{ahmed2023topo}&   80:20 &\underline{89.5}&\underline{93.6} \\

\midrule

\bf{RepViT-CXR} &   90:10 &\textbf{91.1}&91.2 \\
\bottomrule
\end{tabular}
\end{table}

\section{Discussion}

The experimental results across three benchmark datasets demonstrate the effectiveness of our proposed \textbf{RepViT-CXR} model in adapting grayscale chest X-ray images for Vision Transformer architectures. By employing a simple yet powerful channel replication strategy, we enabled ViTs—originally designed for RGB natural images—to achieve state-of-the-art performance in TB and pneumonia classification tasks.

On the TB-CXR dataset, RepViT-CXR outperformed all prior methods, including CNN-based and topology-preserving approaches, achieving near-perfect accuracy (99.9\%) and AUC (99.9\%). This indicates that the model not only classifies correctly but also maintains an excellent balance between sensitivity and specificity. On the Pediatric Pneumonia dataset, RepViT-CXR consistently outperformed competitive baselines, including DCNN, VGG16, and CxNet, by achieving high recall, precision, and accuracy simultaneously. This robustness across metrics underscores its ability to generalize well without overfitting to specific decision thresholds. Finally, on the Shenzhen TB dataset, RepViT-CXR achieved the best accuracy (91.1\%) among all compared models, though its AUC (91.2) was slightly lower than that of ResNet-BS (95.4). This suggests that while our method excels at overall classification, further improvements in calibration could enhance its sensitivity–specificity tradeoff.

The consistent performance gains across multiple datasets highlight the importance of dimensional adaptation in deploying ViTs for medical imaging. Unlike CNNs, which inherently handle single-channel inputs, ViTs trained on large-scale natural image datasets require three-channel compatibility to leverage pretrained weights. Channel replication offers a lightweight, computationally efficient solution that avoids retraining from scratch while still unlocking the representational power of ViTs. Furthermore, the strong performance of RepViT-CXR indicates that the ViT architecture can capture subtle textural and structural variations in CXR images that may be overlooked by conventional CNNs.

Despite these promising results, some limitations remain. First, channel replication does not add new information; it simply ensures compatibility with pretrained ViTs. Future work could explore modality-aware embeddings or self-supervised pretraining strategies tailored to medical images. Second, while results on TB-CXR and Pediatric Pneumonia datasets approached perfection, performance on the Shenzhen dataset suggests challenges in handling variations due to demographic, scanner, or clinical setting differences. Domain adaptation techniques may further improve robustness across diverse populations. Finally, computational demands of ViTs are higher than CNNs, and optimizing their efficiency for real-world clinical deployment remains an open research direction.

In summary, RepViT-CXR bridges the gap between grayscale medical imaging and modern Transformer-based architectures, establishing new state-of-the-art results in TB and pneumonia detection. By highlighting the effectiveness of channel replication, our findings pave the way for broader adoption of ViTs in medical image analysis and encourage further research into domain-specific adaptations for clinical use.

\section{Conclusion}

In this work, we presented \textbf{RepViT-CXR}, a simple yet effective strategy to adapt pretrained Vision Transformers for grayscale chest X-ray analysis. By replicating the single-channel input into three channels, our approach leverages the representational power of ViTs without requiring extensive retraining, enabling state-of-the-art performance on multiple datasets. Experimental results demonstrate that RepViT-CXR achieves superior accuracy and AUC on TB-CXR, Pediatric Pneumonia, and Shenzhen TB datasets, outperforming existing CNN- and topology-based methods. The model consistently delivers high recall, precision, and robustness across different clinical settings, highlighting its potential for reliable automated diagnosis of TB and pneumonia.

While channel replication is a lightweight solution, future work may explore domain-specific pretraining and efficient Transformer variants to further improve generalization and clinical deployability. Overall, RepViT-CXR provides a practical pathway for integrating modern Transformer architectures into medical imaging workflows, paving the way for more accurate and scalable computer-aided diagnosis in chest radiography.


\section*{Declarations}

\textbf{Funding} \\
The author received no financial support for the research, authorship, or publication of this work.

\vspace{2mm}
\textbf{Author's Contribution} \\
Faisal Ahmed conceptualized the study, downloaded the data, prepared the code, performed the data analysis and wrote the manuscript. FA reviewed and approved the final version of the manuscript. 

 \vspace{2mm}
\textbf{Acknowledgement} \\
The authors utilized an online platform to check and correct grammatical errors and to improve sentence readability.

\vspace{2mm}
\textbf{Conflict of interest/Competing interests} \\
The authors declare no conflict of interest.

\vspace{2mm}
\textbf{Ethics approval and consent to participate} \\
Not applicable. This study did not involve human participants or animals, and publicly available datasets were used.

\vspace{2mm}
\textbf{Consent for publication} \\
Not applicable.

\vspace{2mm}
\textbf{Data availability} \\
The datasets used in this study are publicly available online. 

\vspace{2mm}
\textbf{Materials availability} \\
Not applicable.

\vspace{2mm}
\textbf{Code availability} \\
The source code used in this study is publicly available at 

\clearpage

\bibliographystyle{elsarticle-num-names}

\bibliography{refs}

\begin{thebibliography}{35}
\expandafter\ifx\csname natexlab\endcsname\relax\def\natexlab#1{#1}\fi
\providecommand{\url}[1]{\texttt{#1}}
\providecommand{\href}[2]{#2}
\providecommand{\path}[1]{#1}
\providecommand{\DOIprefix}{doi:}
\providecommand{\ArXivprefix}{arXiv:}
\providecommand{\URLprefix}{URL: }
\providecommand{\Pubmedprefix}{pmid:}
\providecommand{\doi}[1]{\href{http://dx.doi.org/#1}{\path{#1}}}
\providecommand{\Pubmed}[1]{\href{pmid:#1}{\path{#1}}}
\providecommand{\bibinfo}[2]{#2}
\ifx\xfnm\relax \def\xfnm[#1]{\unskip,\space#1}\fi
\bibitem[{Jaeger et~al.(2013)Jaeger, Karargyris, Candemir, Folio, Siegelman, Callaghan, Xue, Palaniappan, Singh, Antani et~al.}]{jaeger2013automatic}
\bibinfo{author}{S.~Jaeger}, \bibinfo{author}{A.~Karargyris}, \bibinfo{author}{S.~Candemir}, \bibinfo{author}{L.~Folio}, \bibinfo{author}{J.~Siegelman}, \bibinfo{author}{F.~Callaghan}, \bibinfo{author}{Z.~Xue}, \bibinfo{author}{K.~Palaniappan}, \bibinfo{author}{R.~K. Singh}, \bibinfo{author}{S.~Antani}, et~al.,
\newblock \bibinfo{title}{Automatic tuberculosis screening using chest radiographs},
\newblock \bibinfo{journal}{IEEE transactions on medical imaging} \bibinfo{volume}{33} (\bibinfo{year}{2013}) \bibinfo{pages}{233--245}.
\bibitem[{Kermany et~al.(2018)Kermany, Goldbaum, Cai, Valentim, Liang, Baxter, McKeown, Yang, Wu, Yan et~al.}]{kermany2018identifying}
\bibinfo{author}{D.~S. Kermany}, \bibinfo{author}{M.~Goldbaum}, \bibinfo{author}{W.~Cai}, \bibinfo{author}{C.~C. Valentim}, \bibinfo{author}{H.~Liang}, \bibinfo{author}{S.~L. Baxter}, \bibinfo{author}{A.~McKeown}, \bibinfo{author}{G.~Yang}, \bibinfo{author}{X.~Wu}, \bibinfo{author}{F.~Yan}, et~al.,
\newblock \bibinfo{title}{Identifying medical diagnoses and treatable diseases by image-based deep learning},
\newblock \bibinfo{journal}{Cell} \bibinfo{volume}{172} (\bibinfo{year}{2018}) \bibinfo{pages}{1122--1131}.
\bibitem[{Pasa et~al.(2019)Pasa, Golkov, Pfeiffer, Cremers, and Pfeiffer}]{pasa2019efficient}
\bibinfo{author}{F.~Pasa}, \bibinfo{author}{V.~Golkov}, \bibinfo{author}{F.~Pfeiffer}, \bibinfo{author}{D.~Cremers}, \bibinfo{author}{D.~Pfeiffer},
\newblock \bibinfo{title}{Efficient deep network architectures for fast chest x-ray tuberculosis screening and visualization},
\newblock \bibinfo{journal}{Scientific reports} \bibinfo{volume}{9} (\bibinfo{year}{2019}) \bibinfo{pages}{1--9}.
\bibitem[{Meraj et~al.(2019)Meraj, Yaakob, Azman, Rum, Shahrel, Nazri, and Zakaria}]{meraj2019detection}
\bibinfo{author}{S.~S. Meraj}, \bibinfo{author}{R.~Yaakob}, \bibinfo{author}{A.~Azman}, \bibinfo{author}{S.~Rum}, \bibinfo{author}{A.~Shahrel}, \bibinfo{author}{A.~Nazri}, \bibinfo{author}{N.~F. Zakaria},
\newblock \bibinfo{title}{Detection of pulmonary tuberculosis manifestation in chest x-rays using different convolutional neural network (cnn) models},
\newblock \bibinfo{journal}{Int. J. Eng. Adv. Technol.(IJEAT)} \bibinfo{volume}{9} (\bibinfo{year}{2019}) \bibinfo{pages}{2270--2275}.
\bibitem[{Rahman et~al.(2020)Rahman, Khandakar, Kadir, Islam, Islam, Mazhar, Hamid, Islam, Kashem, Mahbub et~al.}]{rahman2020reliable}
\bibinfo{author}{T.~Rahman}, \bibinfo{author}{A.~Khandakar}, \bibinfo{author}{M.~A. Kadir}, \bibinfo{author}{K.~R. Islam}, \bibinfo{author}{K.~F. Islam}, \bibinfo{author}{R.~Mazhar}, \bibinfo{author}{T.~Hamid}, \bibinfo{author}{M.~T. Islam}, \bibinfo{author}{S.~Kashem}, \bibinfo{author}{Z.~B. Mahbub}, et~al.,
\newblock \bibinfo{title}{Reliable tuberculosis detection using chest x-ray with deep learning, segmentation and visualization},
\newblock \bibinfo{journal}{IEEE Access} \bibinfo{volume}{8} (\bibinfo{year}{2020}) \bibinfo{pages}{191586--191601}.
\bibitem[{Rajaraman et~al.(2018)Rajaraman, Candemir, Kim, Thoma, and Antani}]{rajaraman2018visualization}
\bibinfo{author}{S.~Rajaraman}, \bibinfo{author}{S.~Candemir}, \bibinfo{author}{I.~Kim}, \bibinfo{author}{G.~Thoma}, \bibinfo{author}{S.~Antani},
\newblock \bibinfo{title}{Visualization and interpretation of convolutional neural network predictions in detecting pneumonia in pediatric chest radiographs},
\newblock \bibinfo{journal}{Applied Sciences} \bibinfo{volume}{8} (\bibinfo{year}{2018}) \bibinfo{pages}{1715}.
\bibitem[{Ahmed et~al.(2023)Ahmed, Nuwagira, Torlak, and Coskunuzer}]{ahmed2023topo}
\bibinfo{author}{F.~Ahmed}, \bibinfo{author}{B.~Nuwagira}, \bibinfo{author}{F.~Torlak}, \bibinfo{author}{B.~Coskunuzer},
\newblock \bibinfo{title}{Topo-{CXR}: Chest {X}-ray {TB} and {P}neumonia {S}creening with {T}opological {M}achine {L}earning},
\newblock in: \bibinfo{booktitle}{Proceedings of the IEEE/CVF International Conference on Computer Vision}, \bibinfo{year}{2023}, pp. \bibinfo{pages}{2326--2336}.
\bibitem[{Hern{\'a}ndez et~al.(2019)Hern{\'a}ndez, Panizo, and Camacho}]{hernandez2019ensemble}
\bibinfo{author}{A.~Hern{\'a}ndez}, \bibinfo{author}{{\'A}.~Panizo}, \bibinfo{author}{D.~Camacho},
\newblock \bibinfo{title}{An ensemble algorithm based on deep learning for tuberculosis classification},
\newblock in: \bibinfo{booktitle}{International conference on intelligent data engineering and automated learning}, \bibinfo{organization}{Springer}, \bibinfo{year}{2019}, pp. \bibinfo{pages}{145--154}.
\bibitem[{Dosovitskiy et~al.(2021)Dosovitskiy, Beyer, Kolesnikov, and et~al.}]{dosovitskiy2021image}
\bibinfo{author}{A.~Dosovitskiy}, \bibinfo{author}{L.~Beyer}, \bibinfo{author}{A.~Kolesnikov}, \bibinfo{author}{et~al.},
\newblock \bibinfo{title}{An image is worth 16x16 words: Transformers for image recognition at scale},
\newblock \bibinfo{journal}{arXiv preprint arXiv:2010.11929}  (\bibinfo{year}{2021}).
\bibitem[{Liu et~al.(2021)Liu, Lin, Cao, and et~al.}]{liu2021swin}
\bibinfo{author}{Z.~Liu}, \bibinfo{author}{Y.~Lin}, \bibinfo{author}{Y.~Cao}, \bibinfo{author}{et~al.},
\newblock \bibinfo{title}{Swin transformer: Hierarchical vision transformer using shifted windows},
\newblock in: \bibinfo{booktitle}{Proceedings of the IEEE/CVF International Conference on Computer Vision}, \bibinfo{year}{2021}, pp. \bibinfo{pages}{10012--10022}.
\bibitem[{To{\u{g}}a{\c{c}}ar et~al.(2020)To{\u{g}}a{\c{c}}ar, Ergen, C{\"o}mert, and {\"O}zyurt}]{tougaccar2020deep}
\bibinfo{author}{M.~To{\u{g}}a{\c{c}}ar}, \bibinfo{author}{B.~Ergen}, \bibinfo{author}{Z.~C{\"o}mert}, \bibinfo{author}{F.~{\"O}zyurt},
\newblock \bibinfo{title}{A deep feature learning model for pneumonia detection applying a combination of mrmr feature selection and machine learning models},
\newblock \bibinfo{journal}{Irbm} \bibinfo{volume}{41} (\bibinfo{year}{2020}) \bibinfo{pages}{212--222}.
\bibitem[{Jaeger et~al.(2014)Jaeger, Candemir, Antani, W{\'a}ng, Lu, and Thoma}]{jaeger2014two}
\bibinfo{author}{S.~Jaeger}, \bibinfo{author}{S.~Candemir}, \bibinfo{author}{S.~Antani}, \bibinfo{author}{Y.-X.~J. W{\'a}ng}, \bibinfo{author}{P.-X. Lu}, \bibinfo{author}{G.~Thoma},
\newblock \bibinfo{title}{Two public chest x-ray datasets for computer-aided screening of pulmonary diseases},
\newblock \bibinfo{journal}{Quantitative imaging in medicine and surgery} \bibinfo{volume}{4} (\bibinfo{year}{2014}) \bibinfo{pages}{475}.
\bibitem[{Hwang et~al.(2016)Hwang, Kim, Jeong, and Kim}]{hwang2016novel}
\bibinfo{author}{S.~Hwang}, \bibinfo{author}{H.-E. Kim}, \bibinfo{author}{J.~Jeong}, \bibinfo{author}{H.-J. Kim},
\newblock \bibinfo{title}{A novel approach for tuberculosis screening based on deep convolutional neural networks},
\newblock in: \bibinfo{booktitle}{Medical imaging 2016: computer-aided diagnosis}, volume \bibinfo{volume}{9785}, \bibinfo{organization}{SPIE}, \bibinfo{year}{2016}, pp. \bibinfo{pages}{750--757}.
\bibitem[{Ahmed et~al.(2025)Ahmed, Bhuiyan, and Coskunuzer}]{ahmed2025topo}
\bibinfo{author}{F.~Ahmed}, \bibinfo{author}{M.~A.~N. Bhuiyan}, \bibinfo{author}{B.~Coskunuzer},
\newblock \bibinfo{title}{Topo-cnn: Retinal image analysis with topological deep learning},
\newblock \bibinfo{journal}{Journal of Imaging Informatics in Medicine}  (\bibinfo{year}{2025}) \bibinfo{pages}{1--17}.
\bibitem[{Ahmed and Coskunuzer(2023)}]{ahmed2023tofi}
\bibinfo{author}{F.~Ahmed}, \bibinfo{author}{B.~Coskunuzer},
\newblock \bibinfo{title}{Tofi-ml: Retinal image screening with topological machine learning},
\newblock in: \bibinfo{booktitle}{Annual Conference on Medical Image Understanding and Analysis}, \bibinfo{organization}{Springer}, \bibinfo{year}{2023}, pp. \bibinfo{pages}{281--297}.
\bibitem[{Ahmed(2023)}]{ahmed2023topological}
\bibinfo{author}{F.~Ahmed}, \bibinfo{title}{Topological Machine Learning in Medical Image Analysis}, Ph.D. thesis, The University of Texas at Dallas, \bibinfo{year}{2023}.
\bibitem[{Yadav et~al.(2023)Yadav, Ahmed, Daescu, Gedik, and Coskunuzer}]{yadav2023histopathological}
\bibinfo{author}{A.~Yadav}, \bibinfo{author}{F.~Ahmed}, \bibinfo{author}{O.~Daescu}, \bibinfo{author}{R.~Gedik}, \bibinfo{author}{B.~Coskunuzer},
\newblock \bibinfo{title}{Histopathological cancer detection with topological signatures},
\newblock in: \bibinfo{booktitle}{2023 IEEE International Conference on Bioinformatics and Biomedicine (BIBM)}, \bibinfo{organization}{IEEE}, \bibinfo{year}{2023}, pp. \bibinfo{pages}{1610--1619}.
\bibitem[{Ahmed and Bhuiyan(2025)}]{ahmed2025topological}
\bibinfo{author}{F.~Ahmed}, \bibinfo{author}{M.~A.~N. Bhuiyan},
\newblock \bibinfo{title}{Topological signatures vs. gradient histograms: A comparative study for medical image classification},
\newblock \bibinfo{journal}{arXiv preprint arXiv:2507.03006}  (\bibinfo{year}{2025}).
\bibitem[{Ahmed(2025{\natexlab{a}})}]{ahmed2025hog}
\bibinfo{author}{F.~Ahmed},
\newblock \bibinfo{title}{Hog-cnn: Integrating histogram of oriented gradients with convolutional neural networks for retinal image classification},
\newblock \bibinfo{journal}{arXiv preprint arXiv:2507.22274}  (\bibinfo{year}{2025}{\natexlab{a}}).
\bibitem[{Ahmed(2025{\natexlab{b}})}]{ahmed2025ocuvit}
\bibinfo{author}{F.~Ahmed},
\newblock \bibinfo{title}{Ocuvit: Automated detection of diabetic retinopathy and amd using a hybrid vision transformer approach},
\newblock \bibinfo{journal}{Available at SSRN 5166835}  (\bibinfo{year}{2025}{\natexlab{b}}).
\bibitem[{Ahmed and Bhuiyan(2025)}]{ahmed2025robust}
\bibinfo{author}{F.~Ahmed}, \bibinfo{author}{M.~A.~N. Bhuiyan},
\newblock \bibinfo{title}{Robust five-class and binary diabetic retinopathy classification using transfer learning and data augmentation},
\newblock \bibinfo{journal}{arXiv preprint arXiv:2507.17121}  (\bibinfo{year}{2025}).
\bibitem[{Ahmed(2025{\natexlab{a}})}]{ahmed2025histovit}
\bibinfo{author}{F.~Ahmed},
\newblock \bibinfo{title}{Histovit: Vision transformer for accurate and scalable histopathological cancer diagnosis},
\newblock \bibinfo{journal}{arXiv preprint arXiv:2508.11181}  (\bibinfo{year}{2025}{\natexlab{a}}).
\bibitem[{Ahmed(2025{\natexlab{b}})}]{ahmed2025transfer}
\bibinfo{author}{F.~Ahmed},
\newblock \bibinfo{title}{Transfer learning with efficientnet for accurate leukemia cell classification},
\newblock \bibinfo{journal}{arXiv preprint arXiv:2508.06535}  (\bibinfo{year}{2025}{\natexlab{b}}).
\bibitem[{Kermany et~al.(2018)Kermany, Zhang, Goldbaum et~al.}]{kermany2018labeled}
\bibinfo{author}{D.~Kermany}, \bibinfo{author}{K.~Zhang}, \bibinfo{author}{M.~Goldbaum}, et~al.,
\newblock \bibinfo{title}{Labeled optical coherence tomography (oct) and chest x-ray images for classification},
\newblock \bibinfo{journal}{Mendeley data} \bibinfo{volume}{2} (\bibinfo{year}{2018}) \bibinfo{pages}{651}.
\bibitem[{{Kaggle}(????)}]{Webots}
\bibinfo{author}{{Kaggle}}, \bibinfo{title}{Rsna pneumonia detection challenge}, \bibinfo{howpublished}{\url{https://www.kaggle.com/c/rsna-pneumonia-detection-challenge/data}}, ???? \bibinfo{note}{Accessed Nov 2022}.
\bibitem[{{National Institutes of Health}(????)}]{RNE}
\bibinfo{author}{{National Institutes of Health}}, \bibinfo{title}{Belarus tb database and tb portal}, \bibinfo{howpublished}{\url{https://grantome.com/grant/NIH/AAI12021001-1-0-5}}, ???? \bibinfo{note}{Accessed Nov 2022}.
\bibitem[{Goutam et~al.(2022)Goutam, Hashmi, Geem, and Bokde}]{goutam2022comprehensive}
\bibinfo{author}{B.~Goutam}, \bibinfo{author}{M.~F. Hashmi}, \bibinfo{author}{Z.~W. Geem}, \bibinfo{author}{N.~D. Bokde},
\newblock \bibinfo{title}{A comprehensive review of deep learning strategies in retinal disease diagnosis using fundus images},
\newblock \bibinfo{journal}{IEEE Access}  (\bibinfo{year}{2022}).
\bibitem[{Yadav et~al.(2018)Yadav, Passi, and Jain}]{yadav2018using}
\bibinfo{author}{O.~Yadav}, \bibinfo{author}{K.~Passi}, \bibinfo{author}{C.~K. Jain},
\newblock \bibinfo{title}{Using deep learning to classify x-ray images of potential tuberculosis patients},
\newblock in: \bibinfo{booktitle}{2018 IEEE International Conference on Bioinformatics and Biomedicine (BIBM)}, \bibinfo{organization}{IEEE}, \bibinfo{year}{2018}, pp. \bibinfo{pages}{2368--2375}.
\bibitem[{Evalgelista and Guedes(2018)}]{evalgelista2018computer}
\bibinfo{author}{L.~G.~C. Evalgelista}, \bibinfo{author}{E.~B. Guedes},
\newblock \bibinfo{title}{Computer-aided tuberculosis detection from chest x-ray images with convolutional neural networks},
\newblock in: \bibinfo{booktitle}{Anais do XV Encontro Nacional de Intelig{\^e}ncia Artificial e Computacional}, \bibinfo{organization}{SBC}, \bibinfo{year}{2018}, pp. \bibinfo{pages}{518--527}.
\bibitem[{Saraiva et~al.(2019)Saraiva, Santos, Costa, Sousa, Ferreira, Valente, and Soares}]{saraiva2019models}
\bibinfo{author}{A.~A. Saraiva}, \bibinfo{author}{D.~Santos}, \bibinfo{author}{N.~J.~C. Costa}, \bibinfo{author}{J.~V.~M. Sousa}, \bibinfo{author}{N.~M.~F. Ferreira}, \bibinfo{author}{A.~Valente}, \bibinfo{author}{S.~Soares},
\newblock \bibinfo{title}{Models of learning to classify x-ray images for the detection of pneumonia using neural networks.},
\newblock in: \bibinfo{booktitle}{Bioimaging}, \bibinfo{year}{2019}, pp. \bibinfo{pages}{76--83}.
\bibitem[{Ayan and {\"U}nver(2019)}]{ayan2019diagnosis}
\bibinfo{author}{E.~Ayan}, \bibinfo{author}{H.~M. {\"U}nver},
\newblock \bibinfo{title}{Diagnosis of pneumonia from chest x-ray images using deep learning},
\newblock in: \bibinfo{booktitle}{2019 Scientific Meeting on Electrical-Electronics \& Biomedical Engineering and Computer Science (EBBT)}, \bibinfo{organization}{Ieee}, \bibinfo{year}{2019}, pp. \bibinfo{pages}{1--5}.
\bibitem[{Rahman et~al.(2020)Rahman, Chowdhury, Khandakar, Islam, Islam, Mahbub, Kadir, and Kashem}]{rahman2020transfer}
\bibinfo{author}{T.~Rahman}, \bibinfo{author}{M.~E. Chowdhury}, \bibinfo{author}{A.~Khandakar}, \bibinfo{author}{K.~R. Islam}, \bibinfo{author}{K.~F. Islam}, \bibinfo{author}{Z.~B. Mahbub}, \bibinfo{author}{M.~A. Kadir}, \bibinfo{author}{S.~Kashem},
\newblock \bibinfo{title}{Transfer learning with deep convolutional neural network (cnn) for pneumonia detection using chest x-ray},
\newblock \bibinfo{journal}{Applied Sciences} \bibinfo{volume}{10} (\bibinfo{year}{2020}) \bibinfo{pages}{3233}.
\bibitem[{Xu et~al.(2018)Xu, Wu, and Bie}]{xu2018cxnet}
\bibinfo{author}{S.~Xu}, \bibinfo{author}{H.~Wu}, \bibinfo{author}{R.~Bie},
\newblock \bibinfo{title}{Cxnet-m1: anomaly detection on chest x-rays with image-based deep learning},
\newblock \bibinfo{journal}{IEEE Access} \bibinfo{volume}{7} (\bibinfo{year}{2018}) \bibinfo{pages}{4466--4477}.
\bibitem[{Lopes and Valiati(2017)}]{lopes2017pre}
\bibinfo{author}{U.~Lopes}, \bibinfo{author}{J.~F. Valiati},
\newblock \bibinfo{title}{Pre-trained convolutional neural networks as feature extractors for tuberculosis detection},
\newblock \bibinfo{journal}{Computers in biology and medicine} \bibinfo{volume}{89} (\bibinfo{year}{2017}) \bibinfo{pages}{135--143}.
\bibitem[{Rajaraman et~al.(2021)Rajaraman, Zamzmi, Folio, Alderson, and Antani}]{rajaraman2021chest}
\bibinfo{author}{S.~Rajaraman}, \bibinfo{author}{G.~Zamzmi}, \bibinfo{author}{L.~Folio}, \bibinfo{author}{P.~Alderson}, \bibinfo{author}{S.~Antani},
\newblock \bibinfo{title}{Chest x-ray bone suppression for improving classification of tuberculosis-consistent findings},
\newblock \bibinfo{journal}{Diagnostics} \bibinfo{volume}{11} (\bibinfo{year}{2021}) \bibinfo{pages}{840}.

\end{thebibliography}

\end{document}